\pdfoutput=1

\documentclass[letterpaper, 10 pt, conference]{ieeeconf}  

\IEEEoverridecommandlockouts                              

\usepackage{graphicx}
\usepackage{algorithm}   
\usepackage{algorithmic}
\usepackage{wrapfig}
\usepackage{amsfonts}
\usepackage{booktabs}
\usepackage{multirow}
\usepackage{pifont}
\usepackage[table]{xcolor}
\usepackage[most]{tcolorbox}
\usepackage{mathrsfs}
\definecolor{LightGray}{gray}{0.97}
\definecolor{LightBlue}{rgb}{0.97,0.985,1.0}
\definecolor{LightOrange}{rgb}{1.0,0.97,0.94}
\definecolor{LightGreen}{rgb}{0.98,1.0,0.98}
\definecolor{LightPurple}{rgb}{0.96,1.0,1.0}

\usepackage{amsmath,amssymb}
\usepackage{hyperref}
\usepackage{xspace}
\usepackage{threeparttable}

\title{\LARGE \bf Open-Vocabulary Object-Goal Navigation \\ by Generalizing Semantic Mapping with Dense CLIP}

\author{Anonymous}
\author{Meng Wei$^{1,2}$, Chenyang Wan$^{1,3}$, Tai Wang$^{1}$, Yuqiang Yang$^{1}$, Wenzhe Cai$^{1}$, Yilun Chen$^{1}$, \\ Hanqing Wang$^{1}$, Jiangmiao Pang$^{1,\ddagger}$, Xihui Liu$^{2,\ddagger}$
\thanks{$^{1}$Shanghai AI Lab, $^{2}$The University of Hong Kong, $^{3}$Zhejiang University}\thanks{ $^{\ddagger}$Corresponding Authors}}

\begin{document}

\maketitle
\thispagestyle{empty}
\pagestyle{empty}

\begin{abstract}
Object-oriented embodied navigation tasks require agents to locate specific objects, either defined by category or images, in unseen environments. While recent methods have made progress in extending closed-set models to open-vocabulary scenarios with foundation models, they typically rely on training-free large language models (LLMs) or finetuning with end-to-end reinforcement learning (RL). 
However, they face challenges in efficiency (e.g., the overhead and cost of LLM inference) and limited generalization from intensive RL training.
In this paper, we propose OVExp, a training-efficient framework for open-vocabulary exploration. We make the first effort to demonstrate the generalization capabilities of semantic map-based goal prediction networks using Dense CLIP models. 
A major challenge is that preserving both precise point-wise object locations and generalizable visual representations in the semantic map leads to unaffordable training costs.
To address this, we design a Cross-Modal Transfer on Semantic Mapping strategy which adapts an intriguing text-only training and transfer to multi-model semantic mapping and goals in test-time. 
Despite relying on text-based spatial layouts with limited objects, OVExp demonstrates robust generalization to unseen targets on established ObjectNav benchmarks.
\end{abstract} 
\section{Introduction}
\label{sec:intro}

Object-oriented visual navigation tasks require an embodied agent to locate and reach object goals in unseen environments.
The goal can be specified either by language prompts that describe the object category~\cite{batra2020objectnav} or by an image of the target object~\cite{krantz2022instance}.
Since the environment is unknown, agents must explore efficiently before the target becomes visible, which involves reasoning about room layouts, object placements, and inter-object relations.
Although previous methods~\cite{yang2018visual, ye2021auxiliary, semexp, pirlnav, poni, peanut} have made strides in efficient exploration, they are inherently limited to handling only a fixed set of object goals seen during training.
Recent works~\cite{esc, l3mvn, voronav, dragon, guess, lzson} attempt to overcome this by leveraging commonsense knowledge from LLMs in a training-free manner, but the lack of grounding in embodied navigation leads to suboptimal performance.

\textit{Extending models to open-vocabulary scenarios with limited task-specific data} has made great progress in computer vision ~\cite{minderer2024scaling, zhao2022exploiting, zhou2022detecting, Wu_2023_CVPR, ghiasi2022scaling, xu2022simple, lueddecke22_cvpr, cho2024catseg}, leveraging pretrained image-text representation from foundation models like CLIP~\cite{clip}. However, applying it to open-vocabulary exploration policies proves to be non-trivial.
Existing attempts, like EmbodiedCLIP~\cite{embclip} and ZSON~\cite{zson}, integrate CLIP into end-to-end Reinforcement Learning (RL)-based policies but suffer from poor generalization due to the low sample efficiency and sparse rewards. 
In contrast to the intensive RL training, recent modular methods~\cite{poni, peanut} rely on object semantic maps and directly train a goal prediction network using supervised learning.
Despite achieving better training efficiency and performance, it remains unexplored how to enable open vocabulary goal prediction in this paradigm. 

\begin{figure}[t]
    \centering
    \includegraphics[width=1.05\linewidth]{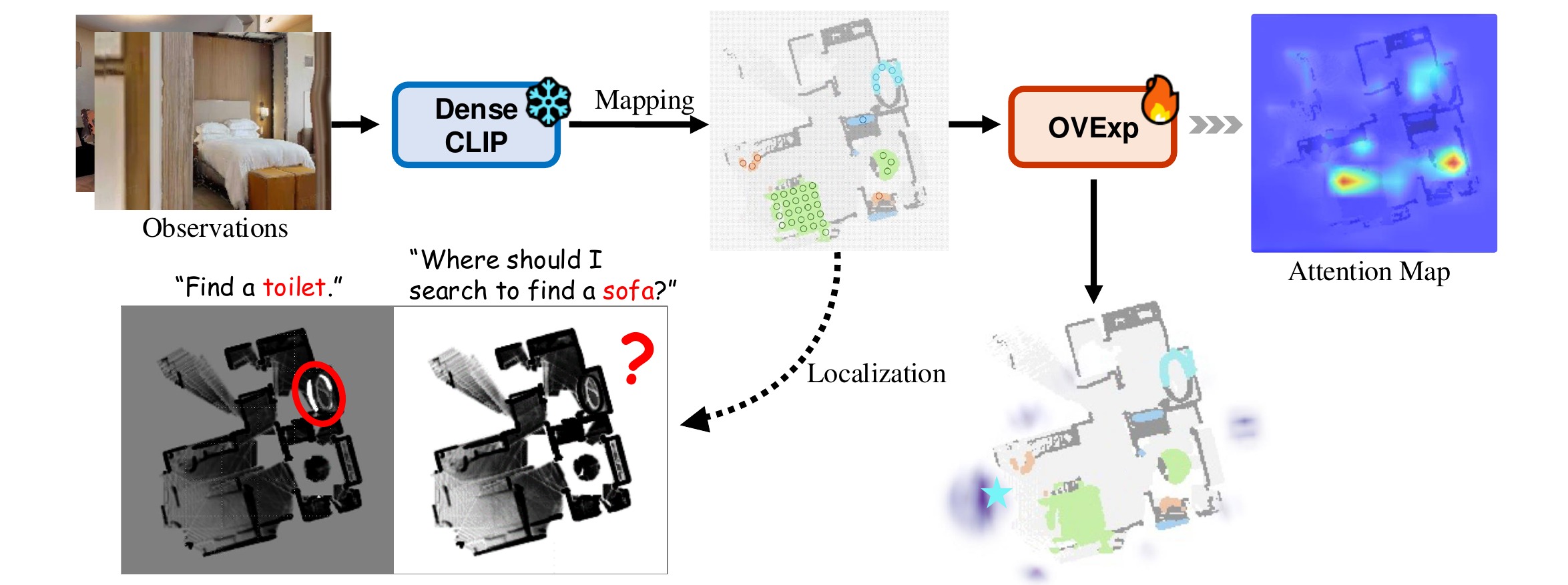}
    \caption{Although VLMaps can localize objects precisely in explored areas, object relationships are not explicitly encoded when querying for unmapped objects. Through fine-tuning, OVExp effectively learns the spatial layout and infers the target's position in unexplored areas.}
    \label{fig:teaser}
    \vspace{-4mm}
\end{figure}
Recently, some methods~\cite{vlmaps, conceptfusion} propose to fuse pixel embeddings from dense CLIP models into VLMaps to support open vocabulary landmark localization (``where it is?'').
In this paper, we take a step further to address the critical aspect of exploration: ``where to search?''. We propose the \emph{OVExp} framework, which is based on goal prediction networks, to integrate the VLMaps into the training loop for open vocabulary exploration.
As shown in Figure~\ref{fig:teaser}, the spatial object relationships are initially not encoded in the map with the exploration-based query. 
But with precise object locations and generalizable representations, OVExp is fine-tuned to understand the spatial layout in explored areas and infer the target position in unexplored areas.

The key strategy in OVExp is the \textit{Cross-Modal Transfer on Semantic Mapping} (CMT-SM). 
Scaling the original VLMaps~\cite{vlmaps, conceptfusion} will create significant storage demands and I/O burden to the training process. 
Hence, we propose to train the OVExp policy in \textit{a text-only way} and directly transfer it to handle vision-based map representations and goals in test-time.
Specifically, we build the low-cost semantic maps with ground-truth layout and encode object labels with CLIP \emph{text} encoder for training and use the well-aligned \emph{visual} features to build the map for inference.   
The frozen map feature is then modulated by a Feature-wise Linear Modulation (FiLM) module conditioned on the goal features.
Finally, the modulated features pass through a few self-attention layers and a transposed convolution model to generate the prediction map.
Notably, the training procedure only needs the geometry layout in the text-form, which indicates an intriguing possibility that we may decouple visual perception from training exploration policies, \emph{e.g.}, we can directly generate text-only layouts instead of scaling up the scene data with realistic textures.

To verify the open-vocabulary exploration capabilities of our framework, we conduct extensive experiments on object-oriented navigation benchmarks, including \textbf{HM3D-ObjectNav}~\cite{habitatchallenge2022}, \textbf{HM3D-InstanceImageNav}~\cite{habitatchallenge2023}, and the open-vocabulary \textbf{HM3D-OVON}~\cite{ovon}. The results demonstrate that:
(1) supervised text-only training provides significantly better performance and efficiency than other training-based models;
(2) with reasonable data collection and training costs, OVExp substantially outperforms training-free methods that rely on unstable and cumbersome LLM interactions (e.g., GPT-2, GPT-3.5);
(3) although MLLMs like GPT-4 achieve competitive results without training, their inference cost is prohibitive;
(4) models trained with text-only input can be effectively adapted to vision-only inference.


\section{Related Work}
\noindent\textbf{Map-based Navigation.}
Map-based representations have proven effective for navigation, enabling efficient path planning with spatial awareness and history information. Common map types include occupancy maps~\cite{activeslam}, topological graphs~\cite{voronav}, semantic maps~\cite{semexp}, and implicit maps~\cite{rim}.
Semantic maps, which incorporate high-level information, allow follow-up works~\cite{poni,peanut} to predict long-term goal probabilities, achieving state-of-the-art performance without reinforcement learning.
Leveraging advances in semantic representation from Large Vision-Language Models (VLMs), VLMaps~\cite{vlmaps} project pixel features from LSeg~\cite{lseg} onto high-dimensional semantic maps, but they require constructing a complete map before open-vocabulary goal indexing.
In contrast, we explore how VLMs can directly empower open-vocabulary exploration.
\textbf{Open Vocabulary Object Navigation.}
Navigating to open-vocabulary object goals is a realistic yet challenging task, prompting the creation of benchmarks~\cite{cow, doze, goat} to advance solutions.
Some methods~\cite{zson} use images as goals, encoded with CLIP to generalize to diverse objects.
A recent trend leverages Large Language Models (LLMs) for training-free exploration~\cite{esc,l3mvn,voronav,dragon,guess}, exploiting commonsense knowledge to guide goal-directed exploration.
However, relying solely on observation images can limit understanding of the full 3D environment, often resulting in suboptimal decisions.
Consequently, effective open-vocabulary exploration policies remain an open challenge.

\textbf{}
\section{Method}
\label{sec:method}
\subsection{Object-Oriented Navigation Task Definitions}
\label{sec:setup}

\noindent\textbf{Object Goal Navigation (ObjectNav)}:
In the ObjectNav task, the embodied agent is required to navigate to an instance of a given object category in unseen environments, such as ``chair'' or ``table''. The agent is equipped with RGB and Depth cameras capturing observation images and is provided with 
the 3-DoF current pose relative to the start position at each timestep $t$. The agent continuously explores its surroundings until it finds an object instance from the target category. The discrete action space $a_t \in \mathcal{A}$ includes \texttt{move\_forward}, \texttt{turn\_left}, \texttt{turn\_right}, \texttt{look\_up}, \texttt{look\_down}, and \texttt{stop}. The navigation episode ends upon execution of the stop action or upon reaching the timestep limit. Success is achieved when the agent predicts a stop action at a location where the distance to the target object is less than 1 meter and the target object is within view.
\\
\noindent\textbf{Instance-Specific Image Goal Navigation (InstanceImageNav)}:
While in the InstanceImageNav task, the embodied agent must navigate to a \textit{specific} object instance depicted in a provided RGB image. 
Compared to the ObjectNav task, InstanceImageNav is more challenging as it demands the agent to precisely identify and locate the sole target instance in the environment, which adds extra complexity to the navigation process.
Despite this complexity, both tasks are object-centric and the evaluation protocol of ObjectNav can be directly applicable to InstanceImageNav.

\begin{figure*}
    \centering
    \includegraphics[width=\linewidth]{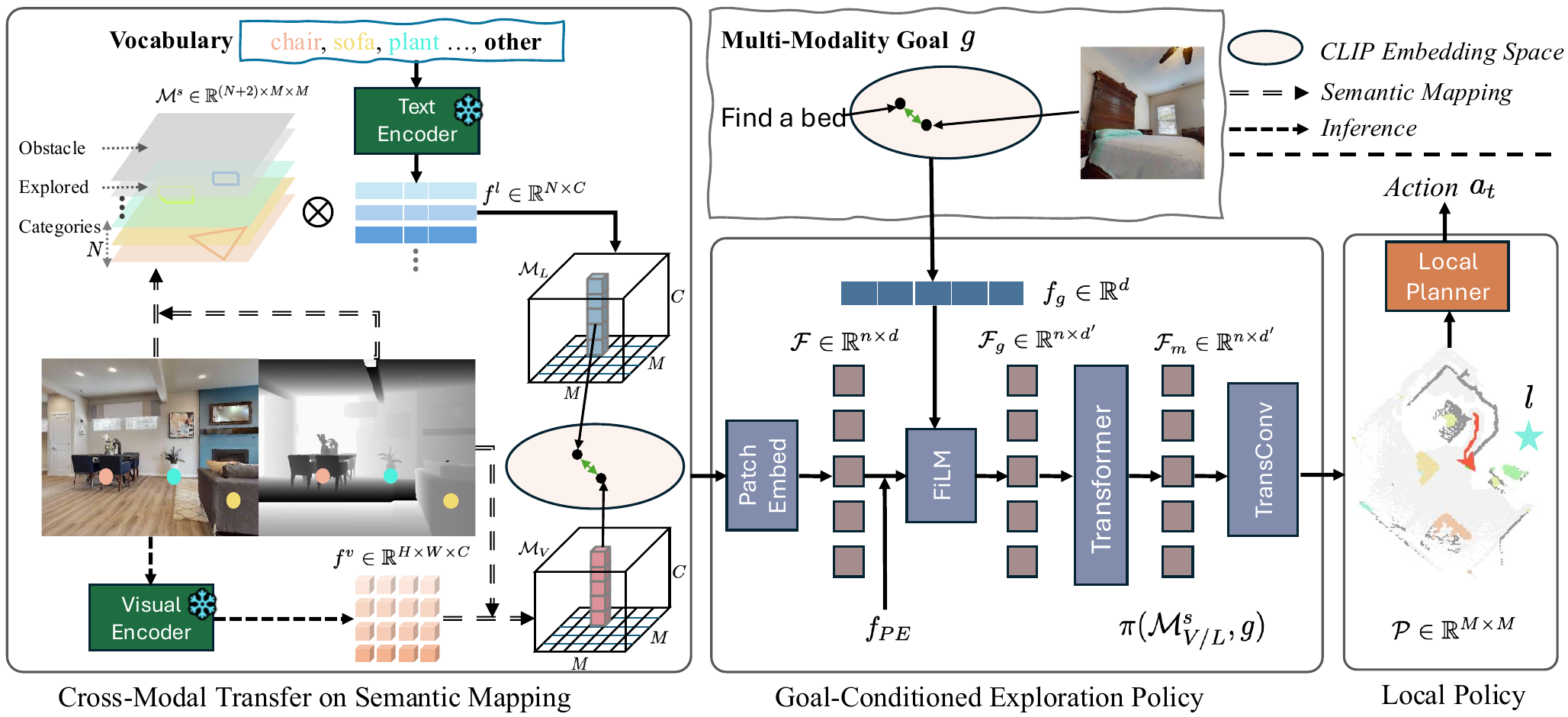}
    \caption{The overall framework of OVExp for open vocabulary object-oriented exploration. 
    OVExp can accept either language-based or vision-based maps as input and accommodates textual and visual object goals.
    For simplicity, the goal identification model is omitted.
    }
    \label{fig:overview}
    \vspace{-4mm}
\end{figure*}
\subsection{Cross-Modal Transfer on Semantic Mapping }
We adopt different types of high-dimensional semantic maps during training and testing phrases, because collecting semantic maps with dense pixel-level visual embeddings from LSeg~\cite{lseg} makes the training impractical (\textit{requiring 1000$\times$ storage demand and 20$\times$ training time than text-only training}).
Therefore, we employ the semantic mapping procedure detailed in Section~\ref{subsec:smap} to construct categorical semantic maps.
We curate a list of 92 object categories from the HM3DSem~\cite{hm3dsem} dataset, which offers abundant pixel-level annotated semantics.
These categorical maps are subsequently transformed into language-based maps using CLIP's text embeddings, detailed in section~\ref{subsec:lmap}.
While in inference, we directly construct maps through vision-based mapping~\ref{subsec:vmap}, as used in VLMaps~\cite{vlmaps}.

\subsubsection{Semantic Mapping}
\label{subsec:smap}
In learning-based goal-oriented exploration, semantic maps~\cite{semexp} have proven to be an effective representation for encoding episodic navigation history.
Semantic maps can capture both geometric priors like spatial layout, obstacles, navigable areas and semantic information such as scene object categories, their locations, and spatial relationships.
To build the semantic map, egocentric visual observations is first segmented into semantic categories using a pre-trained segmentation model. 
Next, a point cloud is extracted from the depth image by back-projection into the 3D world. 
Each point is associated with the corresponding semantic label in the 2D observation image.
The point cloud is then binned into a voxel grid.
Summing along the height dimension, the voxel grid is projected into an egocentric top-down semantic map. 
To be merged with the global map, the egocentric map is finally transformed into the allocentric coordinate system using agent's pose information.
The built semantic map $\mathcal{M}^{s} \in \mathbb{R}^{(N+2) \times M \times M}$ has 2 non-semantic channels (1: obstacle map, 2: explored map) and N semantic channels related to N object categories of interests. 
Each cell within the $M \times M$ grid corresponds to a region of $5 cm \times 5 cm$.

\subsubsection{Language-based Mapping}
\label{subsec:lmap}
To enrich the representation of collected semantic maps, we encode the N semantic channels of $\mathcal{M}^{s}$ with language features derived from CLIP's embedding space.
Formally, we input the text list of N objects $\{O_1, O_2, \ldots, O_N\}$ ($O_N$ represents the background class and use ``other'' as the text) into CLIP's text encoder to produce high-dimensional language-based object features $f^{l} \in \mathbb{R}^ {N \times C}$.
Each semantic channel $\mathcal{M}^{s}_{O_i}  \in \mathbb{R}^{ 1 \times M \times M}$ of the semantic map encodes the confidence score of the object's existence in the grid cells. 
To transform $\mathcal{M}$ into a language-enhanced map representation $\mathcal{M}_{L}$, we compute the weighted sum of the semantic channels and the corresponding language embeddings:
\vspace{-1.5mm}
\begin{equation}
    \mathcal{M}_{L} = \frac{ \textstyle \sum_{i=1}^{N} \mathcal{M}^{s}_{O_i} * f_{O_i}^{l}}{\textstyle \sum_{i=1}^{N}  \mathcal{M}^{s}_{O_i}}
\end{equation}

where $\mathcal{M}_{L} \in \mathbb{R}^ {C \times M \times M}$. 
In $\mathcal{M}_{L}$, the objects are no longer encoded in separate channels; instead, each grid cell contains the averaged language features of all objects present.

\subsubsection{Vision-based Mapping}
\label{subsec:vmap}
During inference, we construct an equivalent vision-based map input with the pretrained image encoder of LSeg, which produces pixel embeddings aligned with their corresponding label embeddings, resembling real-world scene representations. 
The vision-based map construction is a one-stage process.
At each timestep $t$, we extract the dense pixel-wise visual features $f^{v} \in \mathbb{R}^ {H \times W \times C}$ of the observation RGB image $\mathcal{I} \in \mathbb{R}^ {H \times W}$.
The transformation from 2D egocentric pixel representations to a top-down grid map is similar to the semantic mapping process described in Section~\ref{subsec:smap}.
The only difference lies in how the grid map is updated during the online mapping process.
In updating the semantic categorical maps, the maximum object confidence from all timesteps is consistently taken for each grid cell. 
However, the vision-based map $\mathcal{M}_{V}^t$ is updated as follows:
\begin{align}
\begin{split}
    \mathcal{M}_{V}^t\left [ i, j \right ] = \frac{\mathcal{M}_{V}^{t-1}\left [ i, j \right ] \times {\mathcal{N}^{t-1}}\left [ i, j \right ] + m_{v}^t\left [ i, j \right ] }{\mathcal{N}^{t-1}\left [ i, j \right ] + 1}; \\
 \mathcal{N}^{t}\left [ i, j \right ] = \mathcal{N}^{t-1}\left [ i, j \right ] + 1
\end{split}
\end{align}

where $\mathcal{M}_{V} \in \mathbb{R}^{C \times M \times M}$. 
$\left [ i, j \right ]$ represents the locations that will be updated by the incoming map feature $m_{v}^t \in \mathbb{R}^{C \times M \times M}$ which is projected from the current observation feature $f^{v}$.
$\mathcal{N} \in \mathbb{R}^{M \times M}$ records the number of updates in each grid cell over time.


\subsection{Goal-Conditioned Exploration Policy}
\vspace{-2mm}
Given robust high-dimensional semantic maps $\mathcal{M}^{s}_{V/L} \in  \mathbb{R}^{(C+2) \times M \times M}$ (the first two channels represent the obstacle map and explored area), which encode the spatial and semantic information of the environment, and an object goal $g$, our objective is to learn a global policy $\pi$ that outputs the long-term goal location $l$ within the local map:
\begin{equation}
    \pi(\mathcal{M}_{V/L}, g) \to l
\end{equation}
We use $\mathcal{M}^{s}_{L}$ for training and switch to $\mathcal{M^{s}}_{V}$ during inference. 
Following previous map-based methods~\cite{linkmap, peanut}, the policy $\pi$ will output an object probability map and the long-term goal location $l$ is selected with the largest probability.
This policy is flexible to handle goals specified in different modalities, such as a textual goal $g_t$ in ObjectNav and an image goal $g_i$ in InstanceImageNav.

\noindent\textbf{Goal-Conditioned Map Encoder.} The map $\mathcal{M}^{s}_{V/L}$ is first partitioned into non-overlapping patches which are projected into map token embeddings through the patch embed operation.
Then we add learnable positional embedding $f_{PE}$ to the token embeddings and obtain the map features $\mathcal{F}$:
\begin{equation}
    \mathcal{F} = \textsc{PatchEmbed}(\mathcal{M}_{V/L}) + f_{PE}
\end{equation}
where ${\mathcal{F} \in \mathbb{R}^{n \times d}}$, n denotes the number of map tokens and d denotes the hidden size.

To condition the long-term goal prediction on the object goal embedding $f_{g} \in \mathbb{R}^{d}$, which is text or image embedding from CLIP, we employ an efficient Feature-wise Linear Modulation (FiLM)~\cite{film} layer. This layer applies a feature-wise affine transformation to fuse the two sources of features $\mathcal{F}$ and $f_{g}$ , producing a goal-conditioned map feature $\mathcal{F}_{g}$:
\begin{equation}
    \mathcal{F}_{g} = \gamma(f_{g}) \odot h(\mathcal{F}) + \beta(f_{g});
    \mathcal{F}_{m} = \textsc{TRM}(\mathcal{F}_{g})
\end{equation}
where $\mathcal{F}_{g} \in \mathbb{R}^{n \times d'}$ .$\gamma(.)$, $\beta(.)$ and $h(.)$ are three linear transformations.
$\gamma(.)$ and $\beta(.)$ generate the scaling and shifting vectors from $f_{g}$ respectively. $h(.)$ reduces the dimension of $\mathcal{F}$ from $d=512$ to $d'=64$. 
This fusion process effectively integrates the semantic goal information into the map features.
Finally, the encoded map features $\mathcal{F}_{m} \in \mathbb{R}^{n \times d'}$ is produced by feeding $\mathcal{F}_{g}$ to a two-layer transformer to facilitate feature representation learning.

\noindent\textbf{Goal Location Prediction.} To generate the object probability map for selecting the goal location, we design a convolution network as the decoder. The decoder $\mathcal{D}$ consists of one convolution layer and two transposed convolution layers which upsamples the map feature $\mathcal{F}_{m}$ to the original map resolution and generate the goal probability map $\mathcal{P} \in  \mathbb{R}^{M \times M}$.
The location with the highest value in this map is selected as the predicted long-term goal location. Following ~\cite{peanut}, to improve the efficiency of exploration, the final long-term goal location is determined by weighting $\mathcal{P}$ with the geodesic distance to the agent's current location:
\begin{equation}
    \mathcal{P} = \mathcal{D}(f_{g});
    l = arg\max_{i,j}(\mathcal{P}_{ij} \times g_{ij} )
\end{equation}
$(i,j)$ denotes the map index and g is the exponential weight derived from the geodesic distance.
We use the binary cross-entropy loss to train the exploration policy. 
\\
\subsection{Analytical Local Planner}
To reach the predicted goal locations from exploration policy or the identified goal points, we adopt an analytical local planner to translate the long-term goal location into an executable action $a_t \in \mathcal{A}$ as in previous modular map-based methods. 
The Fast Marching Method (FMM) is employed incrementally to calculate the shorted path from the agent's current location to the goal location.
Then a waypoint along this path is selected considering the agent's step distance. 

\section{Experiment}
\subsection{Experimental Setup}
\label{sec:exp_setup}
\noindent\textbf{Training Dataset.}
We generate semantic maps from the HM3DSem~\cite{hm3dsem} dataset using the Habitat~\cite{habitat} simulator. HM3DSem annotates object instances across $216$ 3D scene reconstructions with 1660 raw object names. 
We set a goal-agnostic agent to explore the training scenes from random start locations, allowing it to wander for $500$ steps. During this process, we save the semantic map every $25$ steps. To avoid using maps with minimal unexplored areas, we select only the first half of the saved maps for training.

\noindent\textbf{Evaluation Datasets.}
The evaluation is conducted on the validation sets of three navigation datasets, including one InstanceImageNav dataset and two ObjectNav datasets:

\underline{\texttt{HM3D-ObjectNav}}~\cite{habitatchallenge2022}: Released in the Habitat 2022 challenge, this dataset has 2000 episodes from 20 validation scenes in HM3D, targeting 6 specific goal objects.
We collect the semantic maps based on $86$ objects, \textbf{excluding the 6 goal objects.}

\underline{\texttt{HM3D-OVON}}~\cite{ovon}: An \textbf{open-vocabulary} ObjectNav dataset comprising 379 goal objects across 181 scenes. We collect semantic maps using annotations for 100 objects from the \textbf{Train} split, and evaluate on the \textbf{Val Unseen} split with 49 novel goal objects that \textbf{are absent during training}.

\underline{\texttt{HM3D-InstanceImageNav}}~\cite{habitatchallenge2023}: Released in the Habitat 2023 challenge, this dataset comprises 1000 episodes from 36 validation scenes in HM3D, targeting the instance image goals of 6 objects.\\
\textbf{Evaluation Metrics.}
We evaluate performance with two standard metrics. \textit{\textbf{Success}}: measures the proportion of episodes in which the agent successfully stops near the goal object. \textit{\textbf{SPL}} (Success weighted by Path Length): assesses the efficiency of the navigation by weighting the success rate by agent path length relative to shortest path length.
\subsection{Implementation Details}
\label{sec:imple_details}
We use a global map of size $960 \times 960$ for training, applying random crop operations to $720 \times 720$, along with random flips and random rotations for data augmentation.
We embed the map features with a patch size of $16 \times 16$. Then a 2-layer transformer with the hidden size of $512$ and $8$ attention heads is used to update the map features.
For fusion with FiLM, both the encoded map features and goal embeddings are reduced to a hidden size of $64$.
For the transposed convolutional decoder, the first convolutional layer uses a kernel size of $3$ and a padding size of $1$. The two transposed convolutional layers upsample the feature maps using transposed kernels with a size of $4$.
The prediction model is trained for 20 epochs with a batch size of $8$, requiring $~20$ hours with $8$ NVIDIA V100 GPUs.
We use AdamW optimizer with initial learning rate of $1^{e-4}$ and weight decay of $1^{e-4}$. We use cosine decay.

\definecolor{mygreen}{HTML}{91CCC0}
\definecolor{myorange}{HTML}{F7AC53}
\definecolor{myred}{HTML}{EC6E66}

\begin{table}
\begin{center}
    \setlength{\tabcolsep}{1.4mm}
    \renewcommand{\arraystretch}{1.4}
    \caption{Comparison with state-of-the-art ObjectNav methods on the val set of \textbf{HM3D-ObjectNav}.}
    \begin{tabular}{l|ccccc}
        \toprule
        Method  & Mapping & Trainable & VLM & Success$\uparrow$ & SPL$\uparrow$ \\
        \midrule
        \multicolumn{6}{l}{\textit{Closed-Set Setting}} \\
        SemExp~\cite{semexp} & $\checkmark$ & RL & - & 37.9 & 18.8 \\
        PIRLNav~\cite{pirlnav} & $\times$ & RL\&IL & - & \textbf{61.9} & 27.9 \\
        \rowcolor{myred!18} 
        \underline{PEANUT*}~\cite{peanut} & $\checkmark$ & SL & - & 60.5 & \textbf{30.7} \\
        \rowcolor{myred!18} 
        OVExp* & $\times$ & SL & - & 60.6 & 29.7 \\

        \midrule
        \multicolumn{6}{l}{\textit{Open-Vocabulary Setting}} \\
        CoW~\cite{cow} & $\checkmark$ & $\times$ & CLIP & 32.0 & 18.1 \\
        ZSON~\cite{zson} & $\times$ & $\checkmark$ & CLIP & 25.5 & 12.6 \\
        \rowcolor{myorange!15} 
        L3MVN$\dagger$ ~\cite{l3mvn} & $\checkmark$ & $\times$ & GPT-2 & 50.4 & 23.1 \\
        PixelNav~\cite{pixelnav} & $\times$ & $\checkmark$ & GPT-4 & 37.9 & 20.5 \\
        ZSC~\cite{esc} & $\checkmark$ & $\times$ & GPT-3.5 & 39.2 & 22.3 \\
        VoroNav~\cite{voronav} & $\checkmark$ & $\times$ & GPT-3.5 & 42.0 & 26.0 \\
        VLFM+FMM~\cite{vlfm} & $\checkmark$ & $\times$ & BLIP2 & 50.9 & 23.6\\
        GAMap~\cite{gamap} & $\checkmark$ & $\times$ & CLIP & 53.1 & 26.0 \\
        \rowcolor{mygreen!20} 
        InstructNav$\ddagger$~\cite{instructnav} & $\checkmark$ & $\times$ & GPT-4 & 58.5 &  20.9 \\
        \rowcolor{myred!18}         
        OVExp* & $\checkmark$ & $\checkmark$  & LSeg & \textbf{59.7} & \textbf{28.8} \\
        \rowcolor{myorange!15} 

        OVExp$\dagger$ & $\checkmark$ & $\checkmark$  & LSeg & 58.9 & 26.2 \\
        \rowcolor{mygreen!20} 
        OVExp$\ddagger$ & $\checkmark$ & $\checkmark$  & LSeg & 56.3 & 27.9 \\
        \bottomrule
    \end{tabular}
    \begin{tablenotes}
        “\_” indicates results run from their officially released checkpoint. Methods with the same detection module are marked with superscripts: * for PEANUT, $\dagger$ for L3MVN, and $\ddagger$ for InstructNav.
    \end{tablenotes}
    \label{tab:zs_performance}
    \vspace{-5mm}
\end{center}
\end{table}
\begin{table}[]
    \setlength{\tabcolsep}{2.8mm}
    \renewcommand{\arraystretch}{1.4}
    \centering
    \caption{Comparison on the \textbf{Open Vocabulary HM3D-OVON~\cite{ovon}}}
    \begin{tabular}{ccccc}
            \toprule
            Method     &  Policy  &  VLM &   Success$\uparrow$ & SPL$\uparrow$    \cr 
            \midrule
            DAgger &   DA  & SigLIP  & 10.2               &    4.7   \cr
            PPO &   RL  & SigLIP    & 18.6               &    7.5   \cr
            BCRL &   BC+RL   & SigLIP   & 8.0               &    2.8   \cr
            DAgRL &   DA+RL  & SigLIP    & 18.3               &    7.9   \cr
            VLFM & - & BLIP2 & 35.2   &    19.6   \cr
            OVExp &   SL  & LSeg    & \textbf{37.8}          & \textbf{19.6}      \cr
            \bottomrule
    \end{tabular}
    \begin{tablenotes}
        \item We use the val\_unseen split which contains 49 goal objects that are distinct from the 79 objects used during training.
    \end{tablenotes}
    \label{tab:ovon}
\end{table}

\subsection{Open-Vocabulary ObjectNav Performance.}
\noindent\textbf{Settings.} We evaluate OVExp's generalization ability in navigating to unseen goal objects on HM3D-ObjectNav, which has 6 goal objects. 
To ensure OVExp acquires zero experience with these objects, the model is trained by excluding these 6 objects, i.e., their locations are not learned.

\noindent\textbf{Baselines.} We compare OVExp with three types of existing open vocabulary navigation methods:
\begin{itemize}[leftmargin=*]
\setlength\itemsep{0ex}
\item[--]\texttt{Heuristic-based Exploration}: CoW~\cite{cow} combines a goal-agnostic Frontier-Based Exploration (FBE) algorithm with an open vocabulary goal detector. 
\item[--]\texttt{Learning-based Exploration}: ZSON~\cite{zson} trains the policy on extensive image-goal data which is transferred to object-goal navigation by embedding both types of goals in the same CLIP embedding space.
\item[--]\texttt{VLM-based Exploration}: Most existing open vocabulary exploration methods are training-free and based on Large Vision Language Models to extract the prior knowledge about object arrangements. 
L3MVN~\cite{l3mvn}, ESC~\cite{esc}, PixelNav~\cite{pixelnav} and VoroNav~\cite{voronav} convert observation images to captions mainly about object presence, which are then used to score potential waypoints based on LLM's knowledge of relationships between the observed objects and the goal object. VLFM~\cite{vlfm} and InstructNav~\cite{instructnav} compute explicit value map of target object presence by extracting the similarity scores from large vision language models like BLIP2~\cite{blip2} and GPT-4.
\end{itemize}
\noindent\textbf{Performance Analysis.}
Given the variety of object detectors employed in different methods, a comprehensive comparison poses challenges. Hence, as shown in Table~\ref{tab:zs_performance}, we report results of OVExp using three types of goal detectors from open-source projects: PEANUT, L3MVN and InstructNav for a fair comparison. Notably, when using the same goal detector, the open vocabulary performance of OVExp (purely unseen) is even competitive with the closed-set model (all seen).
Moreover, with reasonable finetuning cost, OVExp outperforms L3MVN by $+8.5\%$ in Success and $+3.1\%$ in SPL. 
Although InstructNav achieves better performance in Success rate with powerful but expensive GPT-4 model, we achieves much better performance in SPL $+7.0\%$, demonstrating better efficiency.

To validate OVExp's generalization ability on more unseen objects, we evaluate OVExp on the HM3D-OVON~\cite{ovon} as shown in Table~\ref{tab:ovon}. We compare OVExp with various learning-based methods which are trained with frozen SigLIP~\cite{siglip} vision, text encoders and then tested on unseen objects. By training on a fixed set of objects and generalizing to novel objects, OVExp achieves significant improvement over the Behavioral cloning (BC), DAgger (DA) and Reinforcement learning (RL) policies.
Comparing to the training-free method VLFM, OVExp achieves better Success Rate. 
\begin{table}[t]
    \renewcommand\arraystretch{1.2}
    \centering
    \fontsize{9}{8}\selectfont
    \caption{Comparison of using vision-based or language-based maps during inference on HM3D Val split.}
    \begin{tabular}{ccc}
        \toprule
        {Semantic Map Modality} & Success$\uparrow$     & SPL$\uparrow$         \cr 
        \midrule
        Text (GroundTruth)   & 59.6                  & 28.7                  \cr
        Vision     & \textbf{60.6}         & \textbf{29.7}         \cr
        \bottomrule
    \end{tabular}
    \label{tab:maptype}
\end{table}

\begin{table}[t]
    \renewcommand\arraystretch{1.2}
    \centering
    \fontsize{9}{8}\selectfont
    \caption{Evaluation of the vision-only inference on the val set of HM3D-InstanceImageNav.}
    \begin{tabular}{cccc}
        \toprule
        {Method}         &  Config  &     Success$\uparrow$ & SPL$\uparrow$    \cr 
        \midrule
        Mod-IIN~\cite{mod-iin} &   Stretch             & 56.1               & \textbf{23.3}      \cr
        OVExp    &   Stretch             & \underline{59.7}               & \underline{21.5}     \cr
        OVExp    &   LoCoBot             & \textbf{63.0}               & 20.5      \cr
        \bottomrule
    \end{tabular}
    \label{tab:insimg}
\end{table}

\begin{figure*}[t]
     \centering
     \includegraphics[width=\linewidth]{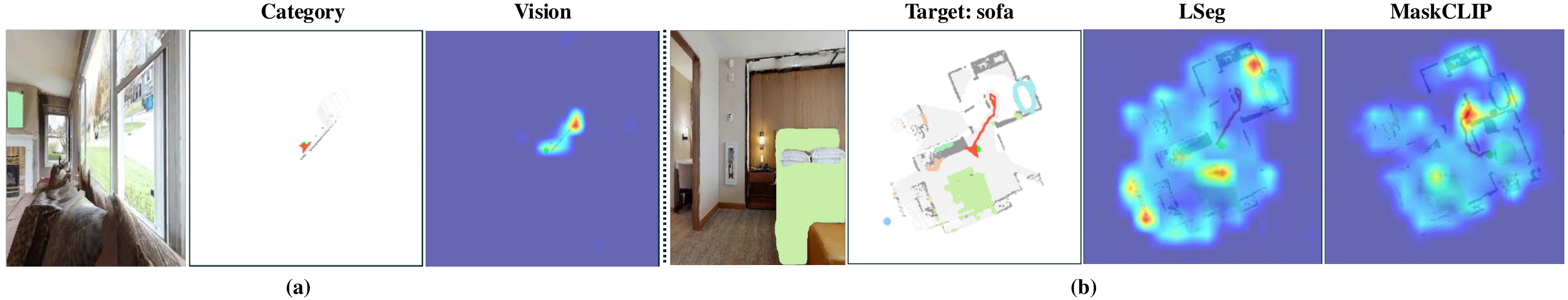}
      \vspace{-4mm}
     \caption{(a) When detection fails, the text-based map lacks context, while the vision-based map remains robust with richer information. (b) In the self-attention layers, LSeg focuses more precisely on relevant objects, while MaskCLIP's attention is more dispersed.}
     \label{fig:attn_vis_compare}
 \end{figure*}
 \begin{figure*}[t]
     \centering
     \includegraphics[width=\linewidth]{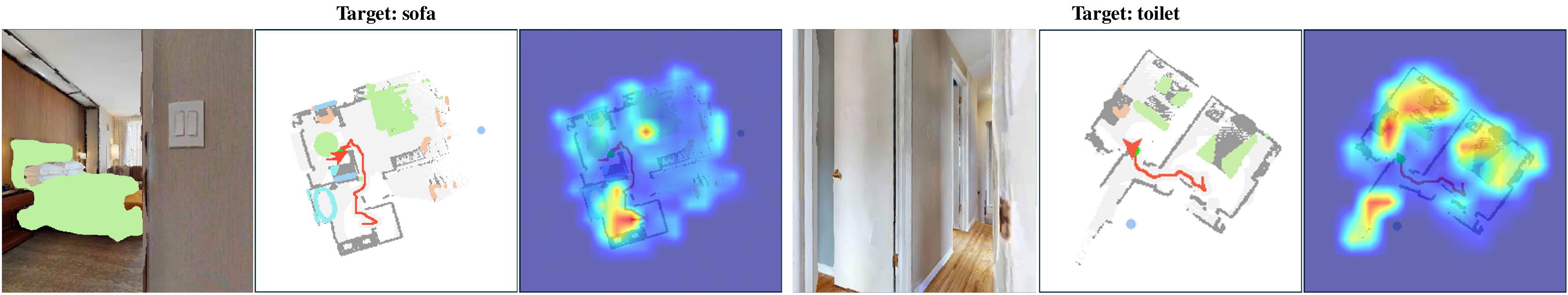}
     \vspace{-4mm}
     \caption{Visualization of attention maps under different goal conditions. When searching for the sofa, the model attends to the bathroom and bed to predict its location. For the toilet, it attends to the chairs, sofa (misdetected) in the living room, bed in the bedroom, and the hallways.}
     \label{fig:attn_vis_lseg}
     
 \end{figure*}


\subsection{Effectiveness of Text-only Training}
\noindent\textbf{Vision-Based Inference Even Surpasses Text-Based Inference.} 
While text and image embeddings from contrastive models like CLIP share some structural alignment, a modality gap still exists. 
Using pixel embeddings instead of text embeddings during inference naturally introduces richer semantic information when constructing the maps. 
To investigate the disparity, we also evaluate the performance using text embedding based semantic maps with the ground truth object annotations. 
The results are reported in Table~\ref{tab:maptype}. 
Surprisingly, ground-truth text-based mapping underperforms vision-based mapping. 
We conjecture that using pixel embeddings during testing introduces variability, which appears as noise in the semantic maps. This noise aids generalization to novel scenes by reducing reliance on overfitted patterns from supervised learning.
Moreover, we find that vision-based mapping is more robust to detection errors. As shown in Figure~\ref{fig:attn_vis_compare} (a), when the sofa is mis-detected, text-based semantic mapping provides no information, whereas vision-based maps can capture this.

\noindent\textbf{Vision-only Inference.}
OVExp is trained in a text-only manner, with both the semantic map and goal features encoded through text. Since text-to-vision transfer is a relatively unexplored paradigm, we conduct an experiment to assess how well OVExp handles this domain shift. Specifically, we perform vision-only inference on the HM3D-InstanceImageNav benchmark to test its adaptability.
In this task, the textual goal embedding is replaced with the image embedding of the goal object, which is more challenging than ObjectNav because the agent must locate a specific object rather than just the first encountered instance.
We compared with the state-of-the-art modular method Mod-IIN~\cite{mod-iin} which proposes a instance re-identification module for goal matching and use FBE as the exploration policy. 
The results are reported in Table~\ref{tab:insimg}.
We observe that OVExp exhibits higher success rates but lower performance on SPL compared to Mod-IIN in both embodiments.
These findings suggest that OVExp, learned for category-based prediction, may not efficiently locate specific instances. 
However, it still achieves a better search strategy than FBE to find the target object eventually.

\subsection{Ablation Study.}
\noindent\textbf{Impact of Semantic Map Resolution} The use of global high-dimensional semantic maps for training incurs the most significant computational cost in our framework. 
 We analyze how different map resolutions can impact effectiveness considering our computational resources. 
 We examine patch sizes of $36, 24, 16$, resulting map sizes of $20 \times 20, 30 \times 30, 45 \times 45$ respectively.
 As shown in table~\ref{tab:mapsize}, using maps with higher resolution contributes to better performance. However, further increasing the map size will result in prohibitively computational cost. Therefore, we adopt the map size of $45 \times 45$.

 \begin{figure*}
    \centering
    \includegraphics[width=\linewidth]{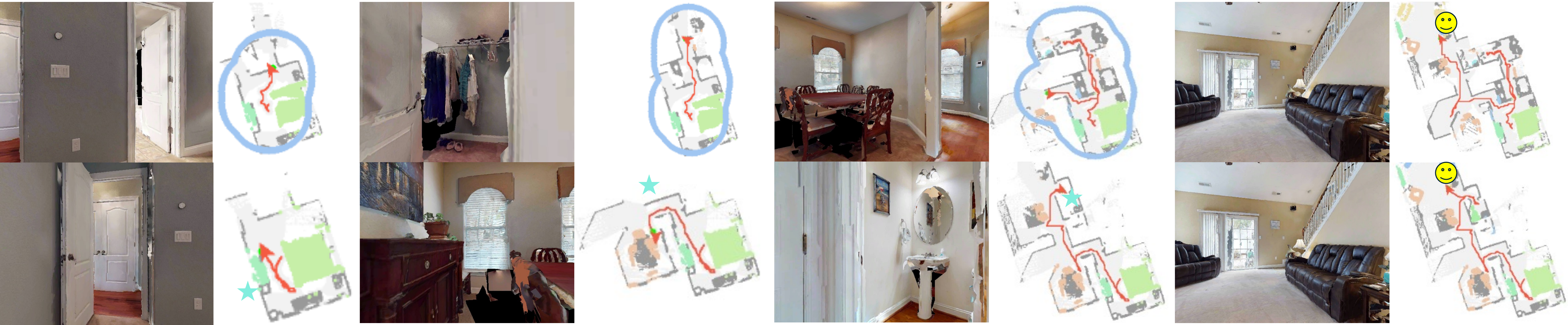}
    \caption{Qualitative results on \texttt{HM3D-ObjectNav}. First row: \emph{FBE}. Second row: \emph{OVExp}.}
    \label{fig:qual_1}
\end{figure*}
\begin{figure*}
    \centering
    \includegraphics[width=\linewidth]{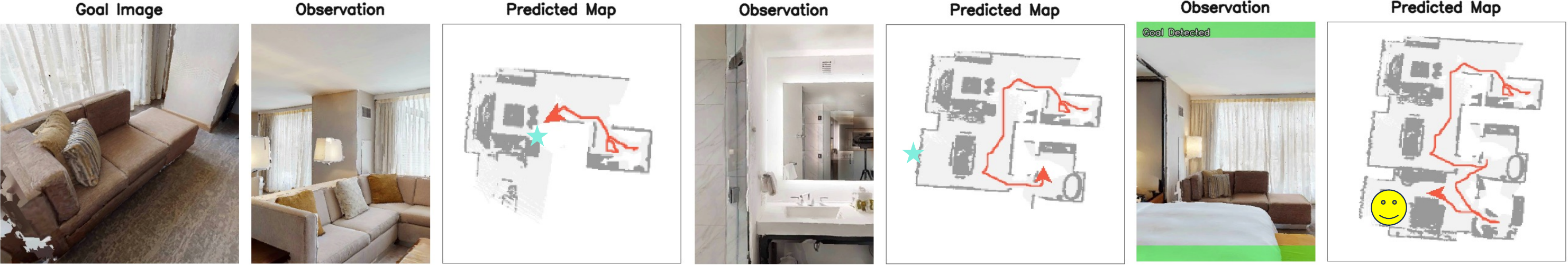}
    \caption{Qualitative results of vision-only inference on \texttt{HM3D-InstanceImageNav}.}
    \label{fig:qual_2}
\end{figure*}

\begin{table}[t]
    \centering
    \fontsize{9}{8}\selectfont
    \caption{Comparison of different map size on HM3D Val split.}
    \begin{tabular}{cccc}
        \toprule
        {Map Size}         &   20 $\times$ 20   &     30 $\times$ 30 & 45 $\times$ 45     \cr 
        \midrule
        Success$\uparrow$  &   56.8             & 58.9               & \textbf{60.6}      \cr
        SPL$\uparrow$      &  26.6              & 27.8               & \textbf{29.7}      \cr
        \bottomrule
    \end{tabular}
    \label{tab:mapsize}
\end{table}
\begin{table}[t]
    \centering
    \fontsize{9}{8}\selectfont
    \caption{Comparison on different types of dense CLIP features.}
    \begin{tabular}{ccc}
        \toprule
        {Dense CLIP Model} & LSeg     & MaskCLIP         \cr 
        \midrule
        Success$\uparrow$   & \textbf{59.4}       & 58.9                  \cr
        SPL$\uparrow$       & \textbf{28.7}       & 28.5        \cr
        \bottomrule
    \end{tabular}
    \label{tab:diffclip}
\end{table}
\noindent\textbf{Impact of Different Dense CLIP Feature.}
We evaluate two types of dense CLIP features—LSeg~\cite{lseg} and MaskCLIP~\cite{maskclip}—to understand the impact on OVExp’s performance. As shown in Table~\ref{tab:diffclip}, using LSeg features achieve better performance than MaskCLIP features.
LSeg, fine-tuned on object segmentation datasets, produce pixel-level embeddings that are closely aligned with object text labels, enabling precise localization in semantic mapping. This fine-grained alignment allow OVExp to identify and map detailed scene elements accurately.
While MaskCLIP adapts CLIP’s global self-attention into a convolutional layer to produce region-based, dense patch features. These features lack the precision needed for scene objects, resulting in less detailed semantic maps.
As shown in Figure~\ref{fig:attn_vis_compare} (b), when predicting the same target, "sofa", LSeg attends to more surrounding context objects compared to MaskCLIP.

\section{Qualitative Results.}
As \emph{OVExp} adopts a trade-off between predicted goal locations and closest locations, we conducted a comparison against the FBE baseline to show the efficiency of \emph{OVExp}. As shown in Figure~\ref{fig:qual_1}, while FBE initially explores the coatroom, leading to inefficient back-and-forth movements, \emph{OVExp} directly navigates out of the bedroom and onwards. 

Furthermore, we offer a qualitative analysis of transferring \emph{OVExp} to the InstanceImageNav task. As \emph{OVExp} is primarily object-oriented, it may lack a nuanced understanding of specific goal details within an image. 
However, it still manages to generate a reasonable exploration path.
As shown in Figure~\ref{fig:qual_2}, \emph{OVExp} first generates a goal location in the living room, where another ``sofa'' is present, before proceeding to the bedroom to locate the goal instance.
And we visualize the attention maps of the transformer to check if the model learn the spatial layout as context as show in Figure~\ref{fig:attn_vis_lseg}. 
We provide more qualitative results in the \textcolor{blue}{\textit{demo video}}.

\section{Real-world deployment.} 
We deploy OVExp on a Turtlebot4 equipped with an RGB-D camera and a 2D LiDAR for odometry and obstacle avoidance. 
The robot continuously sends real-time images, depth data, and poses to a remote server running OVExp. Using these inputs, the server extracts dense visual features, detects target objects, and projects them into a top-down semantic map. OVExp generates open-vocabulary exploration goals within this map and sends them back to the robot, after which Turtlebot4's oracle planner computes an optimal global path to reach the goal.
We evaluate OVExp in an indoor office setting that contains objects unseen during training. 
As shown in the \textcolor{blue}{\textit{demo video}}, OVExp can identify and navigate towards novel goals such as ``sofa’’ or ``printer’’ efficiently. 
This real-world deployment highlights the practicality and scalability of our framework beyond simulation. 
\section{Conclusion}
\label{sec:conclusion}
In this paper, we introduced \emph{OVExp}, a novel modular framework that leverages Dense CLIP models and semantic mapping for open-vocabulary exploration. 
Our approach encodes RGB-D observations with VLMs and project them onto high-dimensional semantic maps.
Specifically, a novel cross-modal transfer on semantic mapping strategy is designed to ensure efficient training with aligned visual-language features.
We train a goal-conditioned exploration policy with only text information and change to the vision-based maps during inference.
The overall design contributes to a flexible framework which can not only process multi-modality maps and goals but also generalize to diverse goals.

\noindent \textbf{Acknowledgements.} This work is supported by Shanghai Artificial Intelligence Laboratory. The research work described in this paper was conducted in the JC STEM Lab of Autonomous Intelligent Systems funded by The Hong Kong Jockey Club Charities Trust.
\bibliographystyle{IEEEtran}
\bibliography{IEEEabrv,root}

\end{document}